\documentclass[conference]{IEEEtran}

\makeatletter

\def\ps@IEEEtitlepagestyle{%
  \def\@oddfoot{\mycopyrightnotice}%
  \def\@evenfoot{}%
}
\def\mycopyrightnotice{%
  {\footnotesize 979-8-3503-
8788-9/24/\$31.00~\copyright~2024 IEEE \hfill}% <--- Change here
  \gdef\mycopyrightnotice{}
}
\usepackage{enumitem}

\usepackage{url}
\usepackage{blindtext}
\usepackage{eso-pic}
\IEEEoverridecommandlockouts
\usepackage{cite}
\usepackage{amsmath,amssymb,amsfonts}
\usepackage{algorithmic}
\usepackage{graphicx}
\usepackage{textcomp}
\usepackage{xcolor}
\usepackage{comment}
\usepackage{booktabs}
\def\BibTeX{{\rm B\kern-.05em{\sc i\kern-.025em b}\kern-.08em
    T\kern-.1667em\lower.7ex\hbox{E}\kern-.125emX}}
    
\usepackage{eso-pic}

\begin{document}
\title{\vspace*{1cm}Harnessing Large Language Models: Fine-tuned BERT for Detecting Charismatic Leadership Tactics in Natural Language}
\author{
\IEEEauthorblockN{Yasser Saeid*}
\IEEEauthorblockA{\textit{Department of Engineering and Economics} \\
\textit{South Westphalia University of Applied Sciences}\\
Meschede, Germany \\
saeid.yasser@fh-swf.de}
\and
\IEEEauthorblockN{Felix Neubürger*}
\IEEEauthorblockA{\textit{Department of Engineering and Economics} \\
\textit{South Westphalia University of Applied Sciences }\\
Meschede, Germany \\
0000-0002-8983-6466}
\and
\IEEEauthorblockN{Stefanie Krügl}
\IEEEauthorblockA{\textit{Department of Engineering and Economics} \\
\textit{South Westphalia University of Applied Sciences }\\
Meschede, Germany \\
0000-0002-2670-2655}
\and
\IEEEauthorblockN{Helena Hüster}
\IEEEauthorblockA{\textit{Department of Engineering and Economics} \\
\textit{South Westphalia University of Applied Sciences }\\
Meschede, Germany \\
0009-0008-9900-7814}
\and
\IEEEauthorblockN{Thomas Kopinski}
\IEEEauthorblockA{\textit{Department of Engineering and Economics} \\
\textit{South Westphalia University of Applied Sciences }\\
Meschede, Germany \\
kopinski.thomas@fh-swf.de}
\and
\IEEEauthorblockN{Ralf Lanwehr}
\IEEEauthorblockA{\textit{Department of Engineering and Economics} \\
\textit{South Westphalia University of Applied Sciences }\\
Meschede, Germany \\
lanwehr.ralf@fh-swf.de}
}

\IEEEoverridecommandlockouts
\IEEEpubid{\makebox[\columnwidth]{978-1-5386-5541-2/18/\$31.00~\copyright2018 IEEE \hfill}
\hspace{\columnsep}\makebox[\columnwidth]{ }}

\maketitle
\IEEEpubidadjcol
\begingroup\renewcommand\thefootnote{*}
\footnotetext{These authors contributed equally to this work}
%\conf{\textit{The 2024 IEEE 3rd Conference on Information Technology and Data, CITDS 2024 \\ 
%26–28 Aug, 2024, Debrecen Hungary }}
\begin{abstract}
This work investigates the identification of Charismatic Leadership Tactics (CLTs) in natural language using a fine-tuned Bidirectional Encoder Representations from Transformers (BERT) model. Based on an own extensive corpus of CLTs generated and curated for this task, our methodology entails training a machine learning model that is capable of accurately identifying the presence of these tactics in natural language. A performance evaluation is conducted to assess the effectiveness of our model in detecting CLTs. We find that the total accuracy over the detection of all CLTs is 98.96\% The results of this study have significant implications for research in psychology and management, offering potential methods to simplify the currently elaborate assessment of charisma in texts.
\end{abstract}


\section{Introduction}
Charismatic leadership is a powerful force influencing human behavior. Consequently, its significance has been a major area of study in psychology, sociology, and management for decades \cite{weber1847}. Charisma, defined as “values-based, symbolic, and emotion-laden leader signaling” \cite[p.~304]{antonakis2016a}, where signals are defined as “things one does that are visible and that are in part designed to communicate” \cite[p.~434]{Spence2002}, is an essential quality that enables leaders to exert their influence effectively. The essence of charismatic leadership lies in the ability of charismatic individuals to build strong connections with their followers. To achieve this connection, they utilize charisma signals to convey their leadership abilities \cite{Akstinaite2024}, express their values \cite{Bastardoz2020,Wilms2022}, and demonstrate their understanding of right and wrong \cite{antonakis2016a}. Leaders such as former U.S. President George W. Bush and former French President François Hollande amplified their charisma following the 9/11 attacks and the Paris terror attacks, indicating that leaders’ charisma signaling significantly increases in response to crises  \cite{Bastardoz2020,Bligh2004}. Previous studies have shown that the effect of charismatic leadership can lead to a range of positive outcomes, such as enhanced follower productivity \cite{Antonakis2022}, increased influence on social media \cite{Tur2022}, and improved adherence to stay-at-home orders during COVID-19, which significantly reduced infection rates and COVID-related death rates \cite{Jensen2023}.  In this context, charismatic leadership tactics (CLTs), which relate to signaling devices for measuring charisma, play a crucial role in understanding and quantifying the phenomenon of charismatic leadership \cite{antonakis2011}. Until now, conducting these studies has been labor-intensive, as all texts for the analysis of charismatic verbal tactics had to be coded manually, because there was no open-source algorithm that could automate this process. However, recent advancements in natural language processing (NLP) techniques, particularly transformer-based models such as BERT \cite{devlin2019bert}, have expanded the possibilities for analyzing the rhetoric of charismatic leadership. By employing these tools, researchers can examine textual data to identify and analyze the linguistic features that underpin CLTs. This paper utilizes this methodology by fine-tuning a BERT model on a corpus of CLTs and applying it to a corpus of generated sentences. The objective is to identify the presence of these tactics within the sentences, to improve the understanding of the linguistic foundations of charisma and to develop analytical tools for studying this phenomenon.
In addition to the primary focus on charismatic leadership, this introduction briefly touches upon related works exploring the interplay of ethical leader signals (ELSs) and emotional intelligence (EI) in the context of leadership ethics. These concepts, closely associated with charismatic leadership, include both verbal and nonverbal cues that communicate ethical values, intentions, and emotional competence among leaders. They have a significant effect on followers' perceptions and organizational outcomes.
One of the most promising approaches for natural language processing tasks is based on transformers, e.g. the BERT model \cite{devlin2019bert}. This model can be fine-tuned on specific corpora to improve its performance on particular tasks, such as sentiment analysis or question answering. The aim of our work is to provide a foundation to facilitate future research on the linguistic means of charismatic leadership by developing tools capable of assisting in analysis and understanding of charismatic leadership. The paper is organized as follows: Section \ref{sec:related} describes related work conducted by researchers that is relevant for this paper. Section \ref{sec:dataset} outlines the data collection process and the construction of a charismatic leadership corpus. Section \ref{sec:setup} elaborates on the methodology employed for fine-tuning the BERT model, while also discussing the experimental outcomes and the effectiveness of our approach. Finally, Section \ref{sec:future} summarizes the findings and proposes potential future research directions in this domain.

\section{Related Works} \label{sec:related}

In recent years, NLP techniques have emerged as promising tools for analyzing and understanding charismatic leadership \cite{choi2017analyzing}. NLP can be used to automatically extract linguistic features from speeches and identify patterns that determine the level of charisma in a text, ultimately aiding in distinguishing charismatic leaders from non-charismatic ones. \cite{choi2017analyzing,tucker2016predicting,gosling2011personality}. For instance, researchers have utilized machine learning algorithms to classify political speeches based on their charismatic content and found that certain linguistic features, such as enthusiasm, optimism, and rhetorical devices, are more prevalent in charismatic speeches \cite{tucker2016predicting}. Another study employed deep learning techniques to analyze the emotional tone of political speeches and demonstrated that charismatic leaders tend to use more positive emotions and fewer negative emotions than non-charismatic leaders \cite{gosling2011personality}.
The advent of transformer-based models, such as BERT, has further advanced the field of NLP by providing robust representations of text data that can capture subtle nuances in language usage \cite{devlin2019bert,liu2020using}. In particular, researchers have fine-tuned pre-trained language models like BERT to classify texts according to their sentiment, emotion, and persuasive appeal \cite{liu2020using}. Building upon this work, we aim to develop a BERT-based model that can accurately detect CLTs in sentences modeled after gubernatorial speeches. Our approach will leverage a large corpus of AI generated sentences modelled after gubernatorial speeches. By training our model on this dataset, we hope to provide insights into how charismatic leadership manifests in spoken language and whether such manifestations vary across different contexts or leader types.
%ethical leadership
In the exploration of ethical leadership, recent literature has critically examined prevailing concepts and advanced new perspectives to enhance our understanding of leadership behaviors. Two notable contributions in this domain shed light on ELS, addressing inherent limitations and proposing innovative frameworks.
A previous study critically evaluated the current landscape of ethical leadership \cite{banks2021ethical}. By identifying conceptual conflation and a dearth of knowledge regarding causes and consequences, the authors propose a refined conceptualization based on signaling theory. Ethical leadership behavior is defined as the signaling behavior of leaders towards stakeholders, which includes the enactment of prosocial values and the expression of moral emotions \cite{banks2021ethical}. This work lays the groundwork for a clearer and more precise understanding of ethical leadership, urging future investigations into theoretical models and methodological advancements.
Complementing this, Banks et al. \cite{banks2022triangulation} used a multifaceted approach to delve into ELSs through signaling theory. Utilizing a combination of constant comparative analysis, preregistered experiments, and data science techniques, the research identifies verbal ELSs associated with emotions such as righteous anger and pride. The study demonstrates the impact of ELSs on evaluations of ethical leadership and delves into their influence on countering counterproductive behavior, enhancing task performance, and encouraging extra role behavior. A noteworthy contribution is the introduction of DeepEthics, a machine learning algorithm designed to automatically score texts for ELSs, offering a practical tool for future research endeavors.
These works collectively emphasize the importance of ELSs in various contexts, including the COVID-19 pandemic and social justice movements. As \cite{antonakis2016a} in their work on charismatic leadership, they argue for a nuanced distinction between leader behaviors and subjective follower evaluations, urging scholars to adopt clearer conceptualizations and innovative methodologies in the ongoing exploration of ELSs. As the field progresses, these contributions serve as valuable guides for researchers seeking to deepen their understanding of the complex dynamics between leaders and their stakeholders in the realm of ethical leadership.
\section{Dataset description} \label{sec:dataset}
In this section we describe the contents of the dataset and explain its generation by using Large Language Models (LLMs)
\subsection{Content of the dataset}
Charismatic leaders often achieve significantly better objective and subjective outcomes in their leadership roles \cite{antonakis2011}. For this reason, most people mention charismatic leaders like Gandhi, Martin Luther King Jr., or Barack Obama when asked about exemplary leaders.
In this signaling process, the charismatic effect is operationalized through twelve CLTs, which consist of three major components: frame, delivery and substance \cite{antonakis2011}. Framing CLTs include the use of metaphors, stories, anecdotes, rhetorical questions, contrasts, and three-part lists, all demonstrating an individual's ability to persuade and effectively utilize figures of speech \cite{antonakis2011}. Charismatic leaders also convey their message through nonverbal behaviors, the delivery CLTs, which are the physical presentation of the signal accompanying the verbal message. Delivery CLTs encompass body gestures, facial expressions, and an animated tone of voice \cite{antonakis2011}. The substance CLTs form the foundation of charismatic leadership. They establish a connection between the leader's message and the followers' needs while clarifying the leader's goals and motivations. This clarity enables followers to trust these powerful signals as they exert effort. Substance CLTs are moral convictions, the expression of  the sentiments of the collective, setting high and ambitious goals, and the leaders’ confidence in the achievement of these goals \cite{antonakis2011}.
Taken together, CLTs enable leaders to signal their ability and willingness to lead and convince individuals to follow them willingly. Since charisma is a strong predictor of positive leadership outcomes, signaling charisma can evoke followers' previous positive experiences with charismatic leaders.
\subsection{Dataset generation using Large Language Models}
We aim at creating a comprehensive dataset for assessing the effectiveness of CLTs and therefore make us of the capabilities of the ChatGPT API \cite{openai_chatgpt}. The resulting dataset, consisting of around 10,000 samples for each tactic, comprises unique subsets aligned with various charismatic strategies. This compilation encompasses a variety of prompts and scenarios, guaranteeing a diverse and well-balanced representation.
\subsection*{Example Prompt}
Below an example prompt for the tactic: setting ambituous goals is displayed. To align the output to expected outputs we are using few-shot prompting by submitting simple examples to the model. The expectation is that the generative model understands the task better than from a general prompt. 

\noindent\fbox{
    \parbox{0.9\columnwidth}{%
        Generate 1232 individual sentences for the following: Generate  sentences expressing explicit goal-setting for followers that is ambitious, often specific. Present each sentence as it would be spoken during a speech, with a focus on achieving maximum diversity. Don't use highfalutin language, but phrase it the way it would be said in a normal presentation Specifically, ensure significant variation in the opening of each sentence to maintain distinctiveness. Each sentence should be unique and notably different from the others. Number each generated response. Here are some examples: 1 "By 2005, all TVs we sell in Japan will be LCD model" 2. "In 2025, I want every single product in our lineup to be made from fully sustainable materials." 3. "In the next seven years, we're going to lead the market, not just in sales, but in sustainability, inclusivity, and innovation." 4. "In five years, we’ll cut our carbon emissions by 80\%, paving the way for a cleaner, greener planet!" 5. "Our goal for 2028 is to implement cutting-edge technology enabling remote work opportunities for all our employees worldwide!" 6. "By the year 2030, we aim to have our software available and fully functional in 25 different languages, expanding our reach and inclusivity!"
    }%
}

The prompts used for dataset generation for the other tactics can be found in the supplementary material in the GitHub repository available at \url{https://github.com/DataScienceLabFHSWF/CharismaAI}
\subsection*{Data Collection and Curation}

\begin{enumerate}[label=\arabic*.]
  \item \textbf{Prompt Development:} We carefully designed prompts with the intention of generating a diverse range of results for each CLT, comprising the potential variability of each respective tactic. To achieve this, we used a profound understanding of linguistic nuances and the pragmatic elements fundamental to charismatic communication. 
  \item \textbf{Output evaluation:} The evaluation of the generated output was carried out by hand. In a first step the prompts were manually evaluated with a small sample size until the desired output was archieved. After a qualitatively satisfying result was found the prompts were slightly adapted for the ChatGPT API to generate large quantities of sentences.
  
  \item \textbf{Data Generation:} Leveraging the ChatGPT API, we methodically produced text samples aligned with each prompt, maintaining a near equal sample size across charismatic tactics. This iterative process, refined over six months, yielded optimized prompt design and consistent output quality.
  
  \item \textbf{Data Curation:} Rigorous curation of text samples ensured that the generated samples have the target CLTs in them. Manual checks were conducted to ensure adherence to charismatic tactics, grammatical accuracy, and overall consistency with prompt instructions.
\end{enumerate}

\subsection*{LLM Evaluation and Comparison}

To examine the performance of various LLMs in generating charismatic text, we conducted a comparative analysis using the generated data set. LLMs, including LLaMa \cite{touvron2023llama} and others, underwent assessment based on their proficiency in producing text aligning with the targeted charismatic tactics. The results of this comparative study however is out of the scope of this paper and will be published in following works.

\subsection*{Data Organization and Analysis}

\begin{enumerate}[label=\arabic*.]
  \item \textbf{Batching and Organization:} In response to ChatGPT API limitations, we segmented data generation requests into smaller batches. Subsequent organization based on charismatic tactics facilitated streamlined data management and analysis.
  
  \item \textbf{Quality Control:} An exhaustive manual examination of all tactics was carried out to ensure the data's overall quality and accuracy. This  approach not only overcame limitations inherent in the API but also enabled a systematic analysis and effective consolidation of the results.
\end{enumerate}
%Our comprehensive research findings will be published in multiple papers, each focusing on a distinct aspect of the data set creation process, LLM performance evaluation, and the broad implications of our work within the field of natural language processing. These publications will collectively provide a holistic understanding of the challenges and opportunities inherent in building and utilizing datasets for assessing charismatic language generation.
%To minimize labeling errors that could compromise the performance and overall classification capabilities of the trained model, we manually curated the generated sentence samples. We will soon publish the exact error rates and a comprehensive statistical analysis of the generated and curated dataset for reference.
In addition to the generated corpus we also used the VU Amsterdam Metaphor Corpus \cite{VUAMetaphor} to extend our corpus with established linguistic ressources.

\section{Methods and Experimental Setup} \label{sec:setup}% generate and hand curate sentences for every tactic
In the domain of multilabel sentence classification for natural language, the BERT algorithm has demonstrated efficacy owing to its advanced capabilities in NLP. BERT, a transformer-based model, has redefined various NLP tasks, including sentence classification, by leveraging bidirectional context understanding and capturing intricate dependencies within textual data.
The foundational mechanism of BERT lies in its pre-training phase, during which it is exposed to vast amounts of unlabeled text data. Through unsupervised learning, BERT learns to predict masked words within sentences, leveraging bidirectional contextual information. This enables BERT to develop a deep understanding of language nuances, making it adept at capturing the intricacies of context and semantic relationships, forming a robust foundation for subsequent tasks such as sentence classification.
In the context of multilabel sentence classification for CLTs, BERT is applied to represent sentences or phrases as embeddings in a high-dimensional vector space.
%% justify with references?
Through tokenization and contextualized embeddings, BERT transforms each word into a vector representation. These vectors are then combined to create contextualized embeddings for the entire sentence, encapsulating the nuanced meaning and context.
The multilabel nature of the task involves predicting multiple labels or categories for a given sentence, aligning with the diverse dimensions inherent in CLTs. 
%class distribution in data
The code and further materials used to implement our methods and experimental setup is publicly available on GitHub \footnote[1]{\url{https://github.com/DataScienceLabFHSWF/CharismaAI}\label{github}} .
% Explain the Method more technical details How did we finetune the BERT model 
% data split for eval
The data containing in total 124776 sentences is split into training and test data with 80\% used for finetuning the BERT model and 20\% for evaluating the models performance. A further split of the training data into train (90\%) and validation set (10\%) is carried out for the internal validation strategy of the pytorch library \cite{pytorch}.
For the tokenization of the sentences the bert-base-uncased tokenizer from the huggingface library \cite{huggingface} was used. This tokenizer ensures that the BERT model recieevs the correct dimensionality of vector representations for training. As the next preprocessing step padding is used to generate sequences of equal length for the models input. 
We then finetune a pretrained BERT model for sequence classification from the huggingface library to classify given sentences into the different CLTs. The model finetuning is then carried out for 100 epochs using the AdamW \cite{adamw} optimizer from the pytorch library with default parameters and a learning rate of $lr = 5 \cdot 10^{-5}$.
\section{Experimental Results and Discussion}
% proof of concept
% classification report
% confusion matrix
% discussion of "problems"
The multilabel classification task is evaluated with a test dataset that was split  from the corpus. The evaluation results are quantized via standard machine learning metrics like accuracy, precision and f1-score. These scores are weighted with classweights and averaged since we do not have a balanced dataset with equal class distributions. The classification evaluation is shown in Table I. The averaged scores are displayed in Table II. The total accuracy of the classification calculates to $98.96\%$. 
\begin{table}[htbp] \label{tab:results1}
\caption{Classification metrics for the different charismatic leadership tactics. }
\begin{tabular}{@{}llllr@{}}
\toprule
Tactic name                 & precision & recall & f1-score & support \\ \midrule
Metaphor/Simile             & 0.9944    & 0.9872 & 0.9908   & 6710    \\
Rhetorical question         & 0.9906    & 0.9996 & 0.9951   & 2431    \\
Story/Anecdote              & 0.9937    & 0.9825 & 0.9881   & 802     \\ 
Contrast                    & 0.9828    & 0.9888 & 0.9858   & 3291    \\
Lists/Repetitions           & 0.9881    & 0.9901 & 0.9891   & 5778    \\
Sentiment of the collective & 0.9835    & 0.9872 & 0.9854   & 2116    \\
Moral conviction            & 0.9864    & 0.9794 & 0.9829   & 1702    \\
Ambitious goals             & 0.9888    & 0.9981 & 0.9934   & 529     \\
Confidence in goals         & 0.9969    & 0.9987 & 0.9978   & 1597    \\ \bottomrule
\end{tabular}
\end{table}
\begin{table}[htbp] \label{tab:results2}
\caption{Averaged classification metrics.}
\centering
\begin{tabular}{@{}lllll@{}}
\toprule
Averaging technique & precision & recall & f1-score &  \\ \midrule
macro avg           & 0.9895    & 0.9902 & 0.9898   &  \\
weighted avg        & 0.9896    & 0.9896 & 0.9896   &  \\ \bottomrule
\end{tabular}
\end{table}

In addition to the quantitative metrics presented in Table I and Table II, a deeper analysis of the multilabel classification task is performed through the examination of a confusion matrix, as illustrated in Figure \ref{fig:conf_matrix}. The confusion matrix provides a more nuanced understanding of the model's performance, particularly in discerning CLTs that exhibit linguistic similarities or partial ambiguity.
In Figure \ref{fig:conf_matrix}, each row corresponds to the actual class, while each column represents the predicted class. The values in the matrix reveal the frequency with which a specific tactic is correctly identified or misclassified. This visual representation allows us to identify patterns of confusion and gain insights into potential challenges faced by the model in distinguishing between certain CLTs.
The observed errors, where tactics may be misclassified due to linguistic similarities, align with the inherent complexities of language and the subjective nature of CLTs. Such challenges are not unique to automated systems but are also subject to interpretation variations among human annotators, emphasizing the need for expert discussions within the field of CLTs.
\begin{figure}[htbp]
    \centering
    \includegraphics[width=\columnwidth, keepaspectratio]{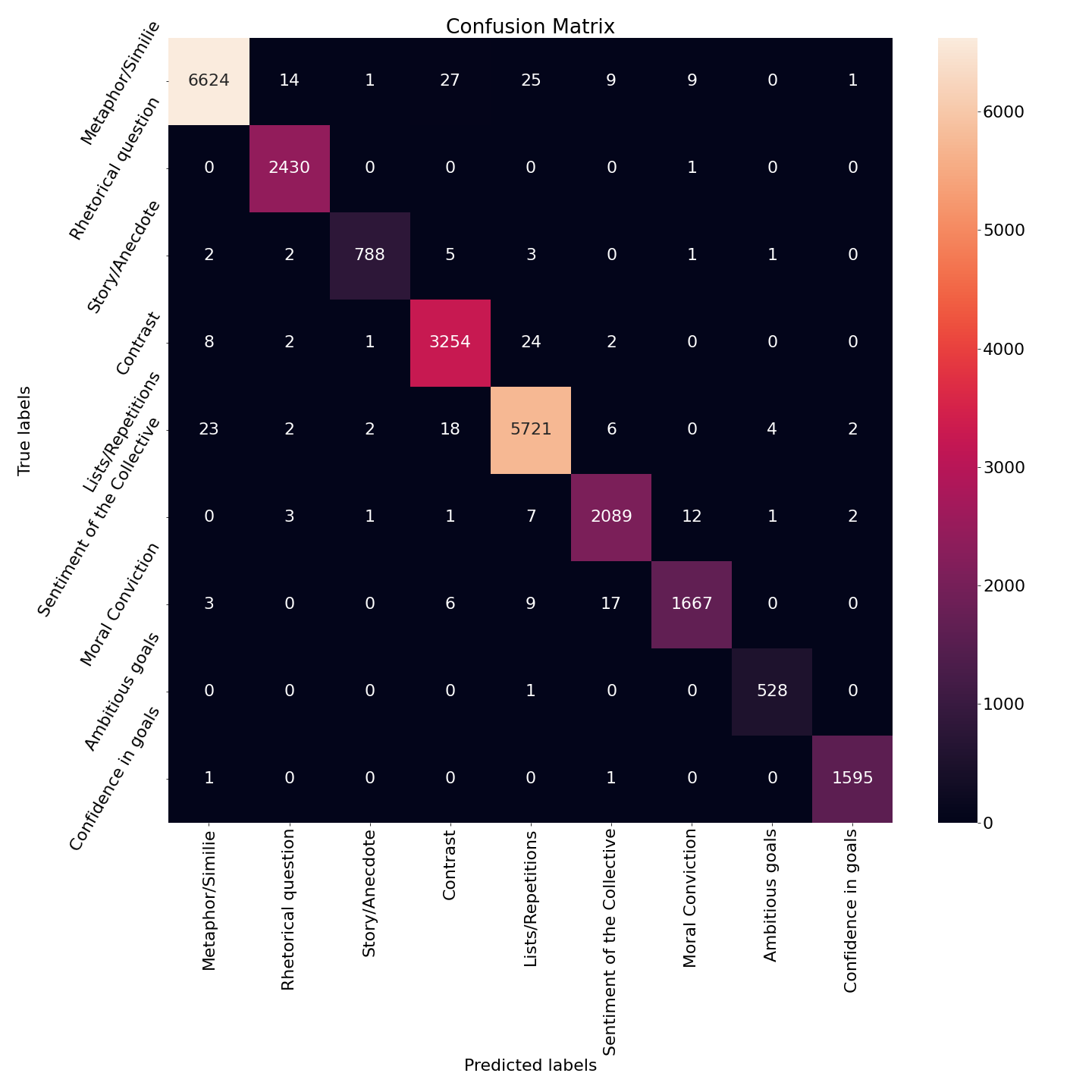}
    \caption{Confusion matrix for the charismatic leadership tactics multilabel classification with BERT.}
    \label{fig:conf_matrix}
\end{figure}
It is important to note that despite these nuanced challenges, the overall performance of the model remains commendable, as evidenced by the high precision, recall, and f1-score values reported in Table I. The precision values, indicating the proportion of correctly identified instances among the predicted positive instances, demonstrate the model's ability to minimize false positives. The recall values, representing the proportion of correctly identified instances among the actual positive instances, highlight the model's effectiveness in capturing the relevant CLTs.
To further validate the robustness of our model, we employ two averaging techniques: macro-average and weighted average. The macro-average computes the unweighted mean of precision, recall, and f1-score, treating each class equally, while the weighted average considers class imbalance by computing the mean with each class's contribution weighted by its prevalence in the dataset. The results, presented in Table II, affirm the model's consistency across different CLTs.
In conclusion, the combination of quantitative metrics, confusion matrix analysis, and averaging techniques provides a comprehensive evaluation of our multilabel classification model. Despite encountering challenges inherent to the linguistic nuances of CLTs, our approach showcases a robust performance that aligns with the intricacies of human interpretation within the field. These findings underscore the model's potential applicability in analyzing and understanding charismatic leadership communication.
% If data was generated, what was the obstacle to creating a balanced dataset?

\section{Future Research} \label{sec:future}
% application to actual gubernatorial speeches speeches (next paper)
% new corpus for public usage (further details in another publication)
% methods for the generation of NLP corpora (using other LLMs?)
% usage and comparison of LLM for sequence classification
Future research in this field holds promising avenues for extending our understanding of CLTs through the lens of NLP. Several potential research directions include:
\begin{itemize}
    \item Application to gubernatorial speeches: \\
    A pertinent next step in this research domain involves the practical application of multilabel sentence classification techniques to (US)-gubernatorial speeches. Analyzing the CLTs employed in real-world political discourse can provide valuable insights into the dynamics of leadership communication. A focused exploration of gubernatorial speeches using advanced NLP models, perhaps extending the current framework to political contexts, could yield nuanced findings.

    \item Using the generated corpus for other NLP research:\\
    Expanding the accessibility of corpora is crucial for advancing research in NLP. The creation of a new corpus, specifically tailored for public usage, presents an opportunity to contribute to the broader research community. Detailed documentation and insights into the construction of such a corpus, along with considerations for ethical and legal aspects, could be outlined in a separate publication. This initiative aims to foster collaboration and enable researchers from diverse backgrounds to explore CLTs using shared datasets.

    \item Methods for the Generation of NLP Corpora (Using Other LLMs):\\
    Exploring alternative methods for generating NLP corpora, particularly leveraging other LLMs, opens avenues for methodological innovation. Investigating the use of different LLMs in corpus generation can shed light on the impact of model architectures on the quality and diversity of corpora. This line of research could provide valuable insights into optimizing corpus creation processes and understanding the role of various LLMs in enhancing the robustness of NLP datasets.

    \item Usage and comparison of LLMs for sequence classification: \\
    The field of NLP constantly evolves with the emergence of new language models. Future research could delve into the usage and comparison of various LLMs for sequence classification tasks, including CLTs. Assessing the performance of different models in capturing contextual nuances and subtle variations in leadership communication can contribute to the refinement of NLP applications in this domain. This comparative analysis could inform researchers on the most suitable LLMs for specific sequence classification tasks related to charismatic leadership.
 
\end{itemize}
These proposed research directions aim to extend the current knowledge base, offering avenues for practical applications, resource expansion, methodological advancements, and model comparisons within the evolving landscape of natural language processing and charismatic leadership analysis.

\bibliographystyle{IEEEtran}
\bibliography{languageresource}

\end{document}